\documentclass[runningheads]{llncs}
\usepackage{graphicx}
\usepackage[caption=false]{subfig}
\usepackage{multirow}
\usepackage{booktabs}
\usepackage{amsmath}
\usepackage{float}
\usepackage{array}
\usepackage{xcolor}
\begin{document}
\title{Talking Detection In Collaborative Learning Environments\thanks{This material is based upon work supported by the National Science Foundation under Grant No.1613637, No.1842220, and No.1949230.}}
%
%
\author{Wenjing Shi\inst{1}
\and
Marios S. Pattichis\inst{1}\and
Sylvia Celed\'on-Pattichis\inst{2}\and
Carlos L\'opezLeiva\inst{2}
}
%
\authorrunning{Wenjing Shi et al.}
%
\institute{Image and Video Processing and Communications Lab\\
	\url{ivpcl.unm.edu}\\ 
	Dept. of Electrical and Computer Engineering\\
	University of New Mexico, United States\and
Dept. of Language, Literacy, and Sociocultural Studies\\
	University of New Mexico, United States.\\
\email{\{wshi, pattichi, sceledon, callopez\}@unm.edu}}
\maketitle              
\begin{abstract}
We study the problem of detecting
   talking activities   
   in collaborative learning videos.
Our approach uses head detection
   and projections of the log-magnitude of optical
   flow vectors to reduce
   the problem to a simple
   classification of small projection
   images without the need for
   training complex, 3-D activity classification systems.
The small projection images are then easily classified
   using a simple majority vote
   of standard classifiers.   
For talking detection, our proposed approach is shown
   to significantly outperform single
   activity systems.
We have an overall accuracy of 59\% compared to
   42\% for Temporal Segment Network (TSN) and 
   45\% for Convolutional 3D (C3D).
In addition, our method is able to detect
   multiple talking instances
   from multiple speakers, while
   also detecting the speakers themselves.
   
\end{abstract}
\section{Introduction}
We study the problem of talking detection in collaborative learning environments.
Here, our ultimate goal is to develop fast
      and reliable methods that can assist
      educational researchers analyze student participation
      in large video datasets.
 
Learning assessment relies heavily
      on the use of audio transcriptions
      that describe the interactions
      between the students and their
      facilitators. 
By identifying the video segments
      where a student is talking,
      educational researchers
      can then further analyze
      the nature of the interactions.
For example, some students may stay quiet.
Others may express themselves throughout
      the lessons.
Ultimately, our computer-based system
      aims at aiding this type of analysis
      by identifying different talking patterns.
However, for the purposes of this paper,
      we will only describe how 
      to reliably detect students talking
      when the camera captures motions over
      their mouths.      
 
\begin{figure}[!t]
\includegraphics[width=\textwidth]{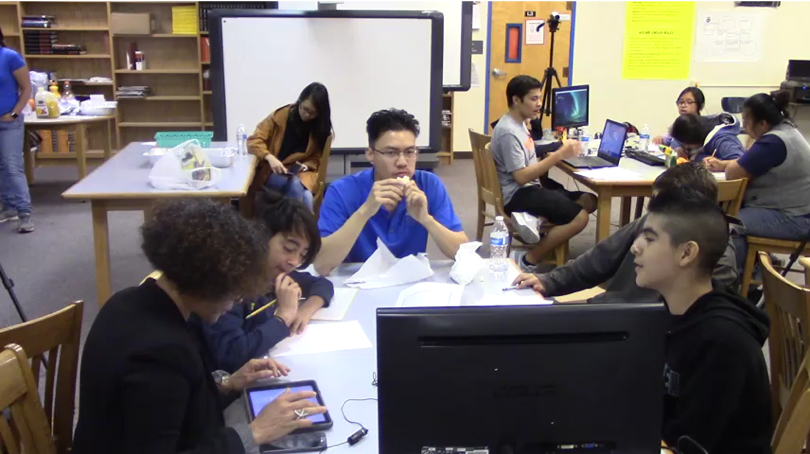}
\caption{A sample that contains multiple challenges for talking detection.} 
\label{fig:AOLMESample}
\end{figure}

We present an example of our collaborative learning environment
   in Fig. \ref{fig:AOLMESample}. 
We are interested in detecting talking activities for the group
   that is closest to the camera.
The students that are farther away appear at a smaller scale and
   need to be rejected from further consideration.
Students appear at different angles to the camera.
Instead of talking, students can also be eating, laughing, or
   yawing, and these activities should not be confused
   with talking (e.g., see eating example in
   Fig.  \ref{fig:AOLMESample}).
In many cases, the mouths may not be visible to the camera.
In such cases, talking detection
   is not possible without processing the audio of the video.

We develop a direct and fast approach to talking detection that
   avoids the need for large training datasets.
First, we detect the heads and faces to include the
   mouth regions.
Then, over the detected head or face regions,
   we compute optical flow vectors and project
   the log-magnitudes of the vectors to generate
   a single region-proposal image over each candidate speaker.
We then use voting from a list of simple classifiers
   to classify each segment as a talking or a non-talking segment.    

Our talking detection research extends prior research by our group.
In \cite{shi2016human}, \cite{Shi2018}
    we introduced the use of multiscale AM-FM decompositions
    to detect student faces and the backs of the heads.
In \cite{tapia2020importance}, the authors demonstrate the importance
    of using the instantaneous phase for face detection.
In \cite{shi2018dynamic},
    we developed methods to identify possible group interactions
    through the use of AM-FM representations. 
In \cite{eilar2016distributed},
    we developed an open-source system for detecting
    writing and typing over cropped video segments.
In \cite{darsey2018hand},
    the author developed a hand movement detection system.
In \cite{jacoby2018context}, we used simple color-based object
    detection followed by classification of optical flow vectors
    to detect writing, talking, and typing over
    a very small number of cropped video segments.

There is also significant human activity detection research
    within the computer vision community.
In \cite{wang2016temporal}, 
    the authors developed the
    Temporal Segment Network (TSN) for video-based activity recognition.
TSN describes a deep learning based approach to detect 
    a diverse range of activities using ConvNets.
In \cite{tran2015learning}, 
    the authors developed the C3D network that
    trains deep 3D convolutional networks
    on a large-scale supervised video dataset
    to detect a diverse range of different activities.
More recently, 
    \cite{tran2018closer} generates a new spatiotemporal convolutional block “R(2+1)D” to train CNNs for activity recognition. 

Our approach avoids the need to train large, deep learning systems
    on human video activity detection.
Our approach is very fast because it reduces talking detection
    to the classification of 
    small proposal regions of the projected motion magnitudes
    over the students' faces or heads.
It is ideally suited for our goal to process over 1,000 hours
    of videos for talking detection.
We also provide comparisons against TSN and C3D to 
    demonstrate that our approach is much more accurate.
    
We organize the rest of the paper into three additional sections.
In section \ref{sec:methodology}, we describe our proposed methodology.
We then provide results in section \ref{sec:results}
   and provide concluding remarks in section \ref{sec:conclusion}. 

\begin{figure}[!t]
	\centering
	\includegraphics[width=\textwidth]{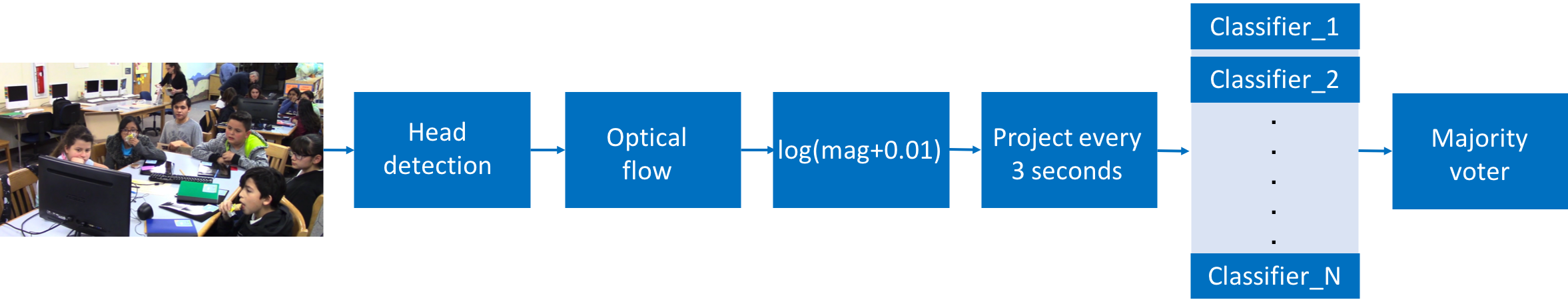}\\
	(a) Talking detection system.\\
	\includegraphics[width=\textwidth]{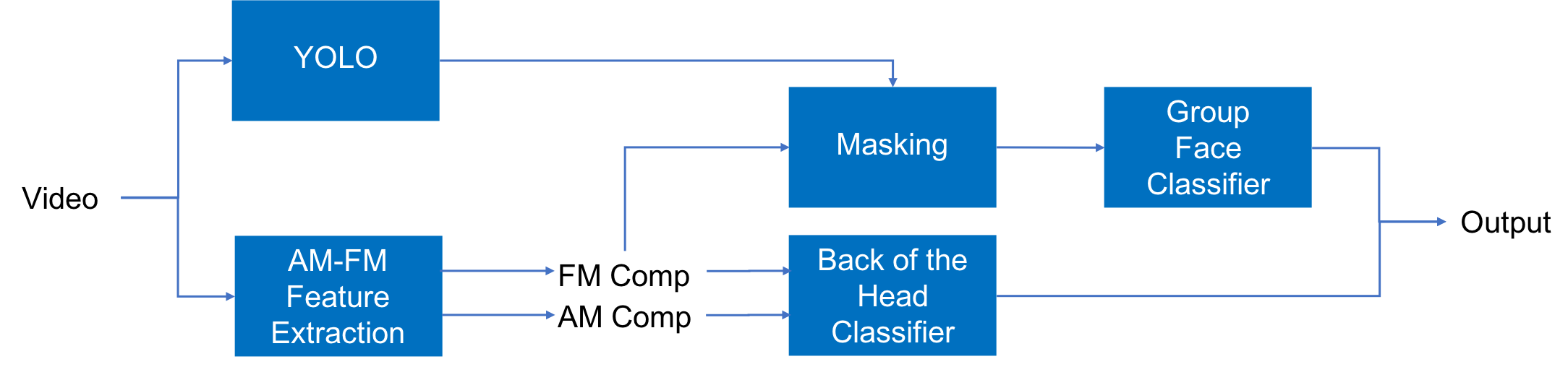}\\
	(b) Head detection system.
	\caption{Group talking detection system.} 
	\label{fig:TalkingDetSystem}
\end{figure}
\section{Methodology}\label{sec:methodology}
We present a system diagram of the entire system in 
   Fig \ref{fig:TalkingDetSystem}(a).
We also include a block diagram for the head detection system
   in Fig \ref{fig:TalkingDetSystem}(b).
In what follows, we summarize the components of each system.    

We use multiple methods to locate both faces and the backs of the heads
   for the head detector.
In the lower branch of 
   Fig \ref{fig:TalkingDetSystem}(b),
   we show how we extract AM-FM features
   using a 54-channel Gabor channel filterbank
   as described in \cite{Shi2018} and \cite{shi2016human}. 
We use AM-FM components to locate the back of the head region. 
For face detection, we use YOLO V3 \cite{redmon2018yolov3}.
We also use FM component classification (LeNet)
   to reject background faces that are characterized by higher
   frequency components since they are farther away from the camera.

\begin{figure}[!t]
	\subfloat{\includegraphics[width=0.155\textwidth]{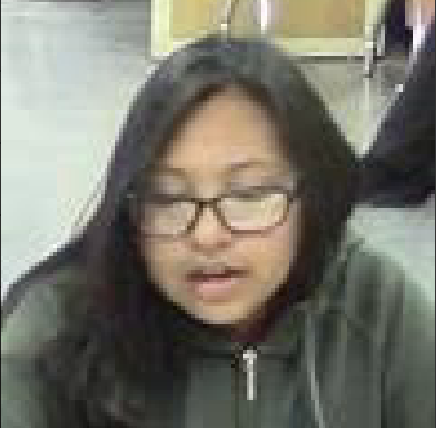}%
	}
	~
	\subfloat{\includegraphics[width=0.155\textwidth]{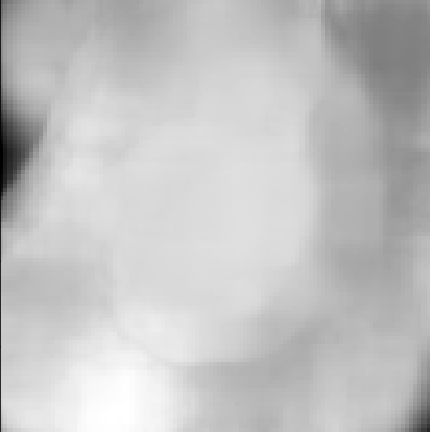}%
	}
	~
	\subfloat{\includegraphics[width=0.155\textwidth]{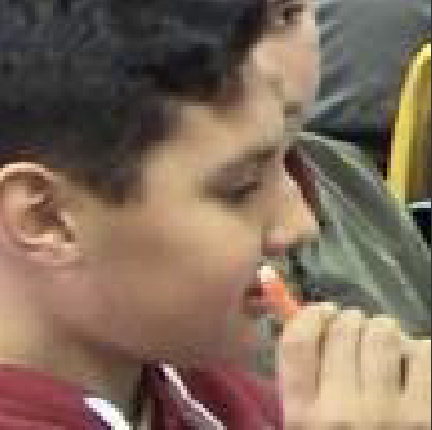}%
	}
	~
	\subfloat{\includegraphics[width=0.155\textwidth]{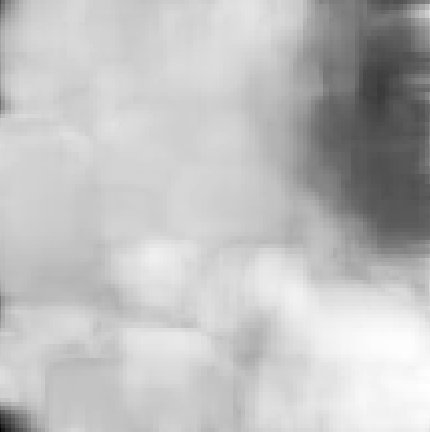}%
	}
	~
	\subfloat{\includegraphics[width=0.155\textwidth]{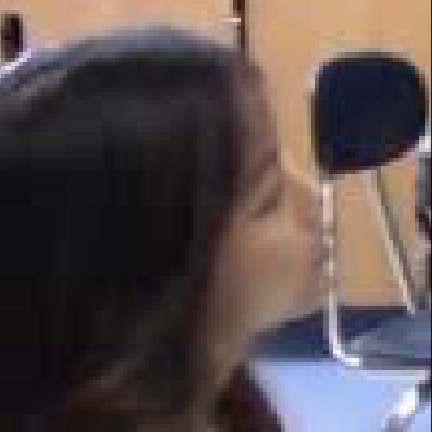}%
	}
	~
	\subfloat{\includegraphics[width=0.155\textwidth]{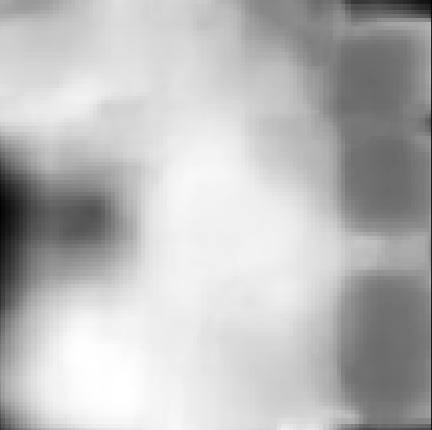}%
	}
	\\
	\subfloat{\includegraphics[width=0.155\textwidth]{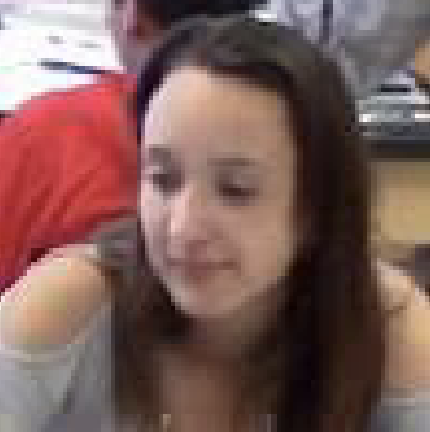}%
	}
	~
	\subfloat{\includegraphics[width=0.155\textwidth]{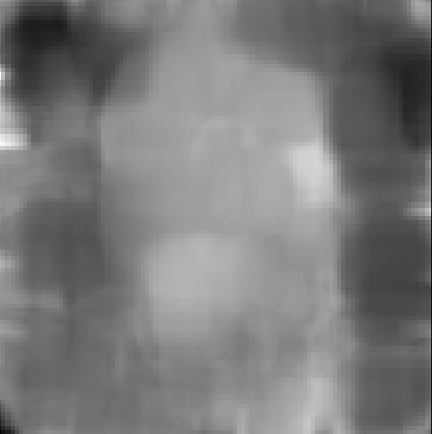}%
	}
	~
	\subfloat{\includegraphics[width=0.155\textwidth]{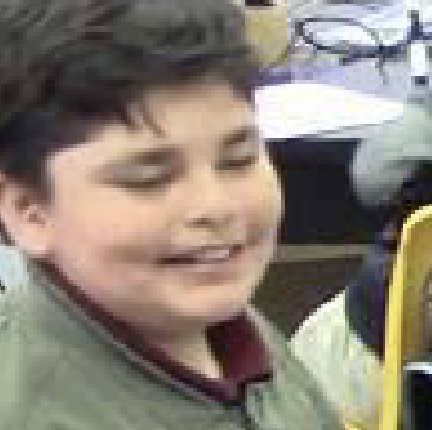}%
	}
	~
	\subfloat{\includegraphics[width=0.155\textwidth]{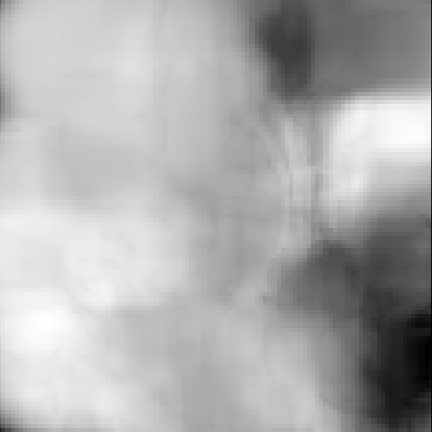}%
	}
	~
	\subfloat{\includegraphics[width=0.155\textwidth]{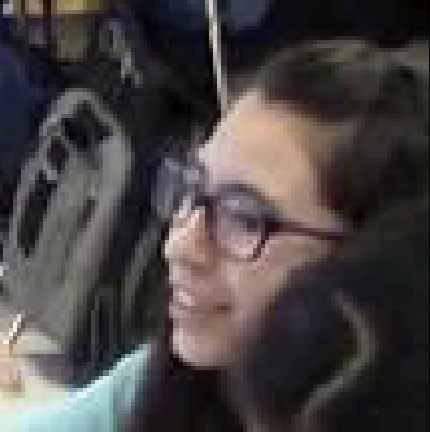}%
	}
	~
	\subfloat{\includegraphics[width=0.155\textwidth]{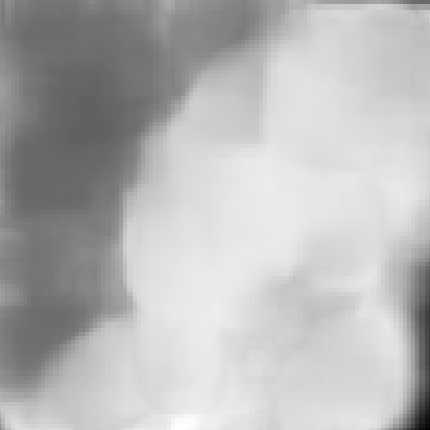}%
	}
	\\~\\
	\subfloat{\includegraphics[width=0.155\textwidth]{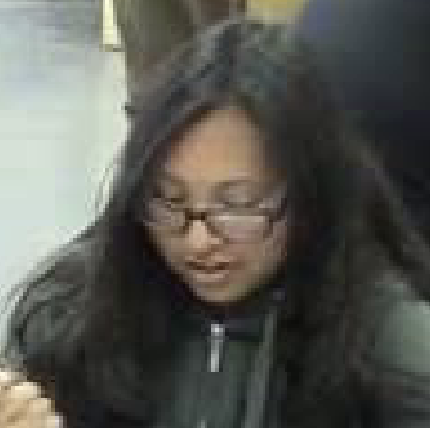}%
	}
	~
	\subfloat{\includegraphics[width=0.155\textwidth]{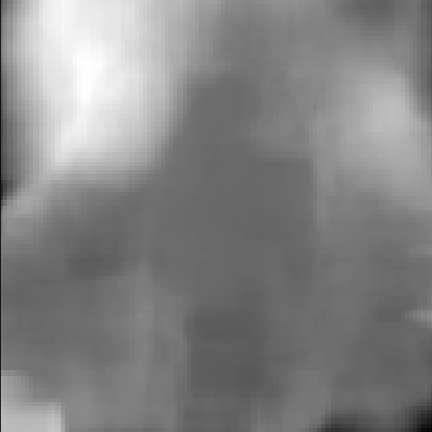}%
	}
	~
	\subfloat{\includegraphics[width=0.155\textwidth]{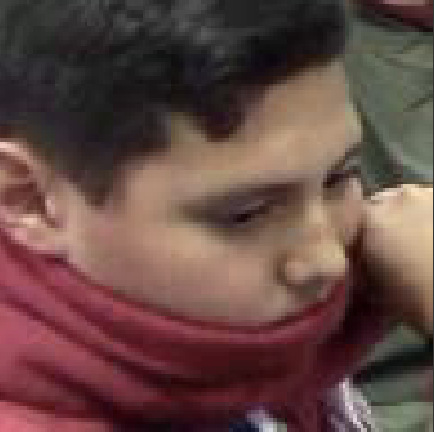}%
	}
	~
	\subfloat{\includegraphics[width=0.155\textwidth]{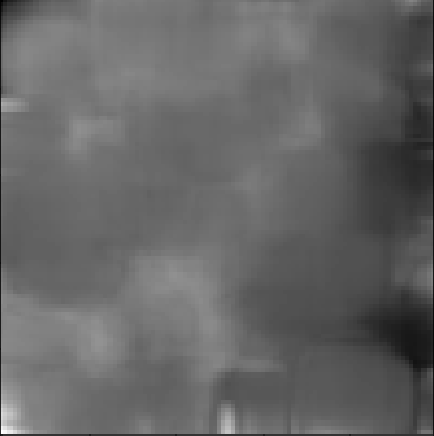}%
	}
	~
	\subfloat{\includegraphics[width=0.155\textwidth]{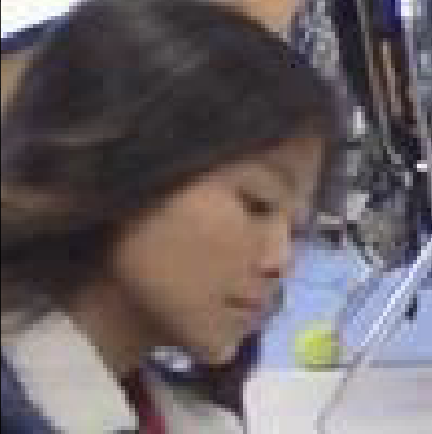}%
	}
	~
	\subfloat{\includegraphics[width=0.155\textwidth]{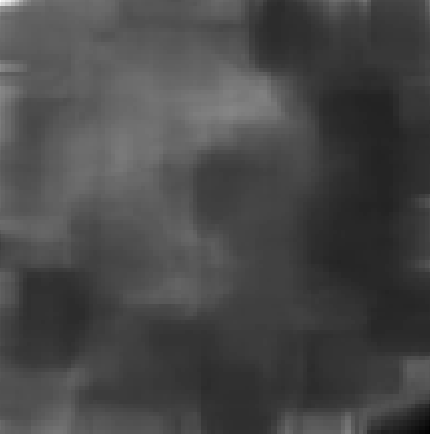}%
	}
	\\
	\subfloat{\includegraphics[width=0.155\textwidth]{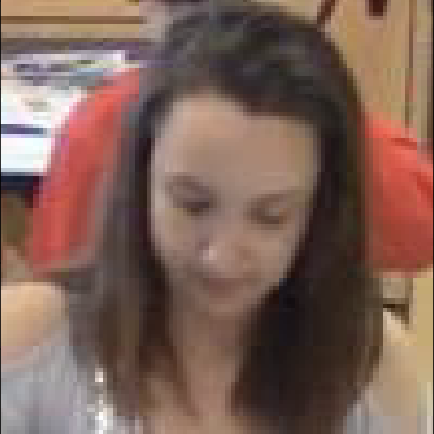}%
	}
	~
	\subfloat{\includegraphics[width=0.155\textwidth]{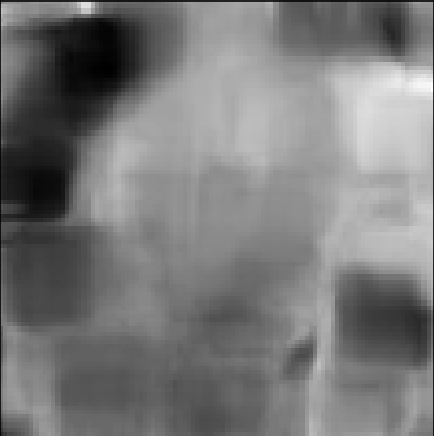}%
	}
	~
	\subfloat{\includegraphics[width=0.155\textwidth]{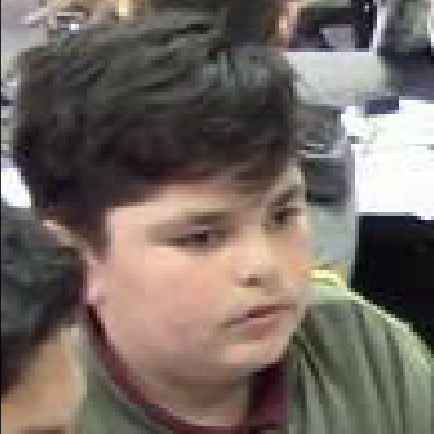}%
	}
	~
	\subfloat{\includegraphics[width=0.155\textwidth]{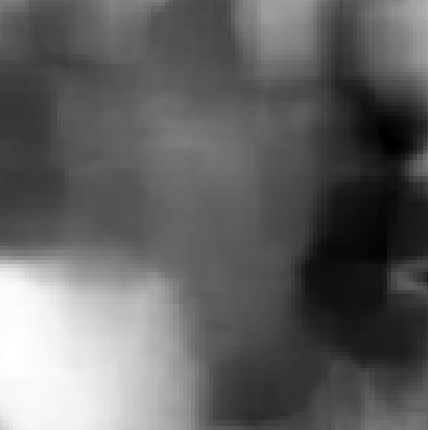}%
	}
	~
	\subfloat{\includegraphics[width=0.155\textwidth]{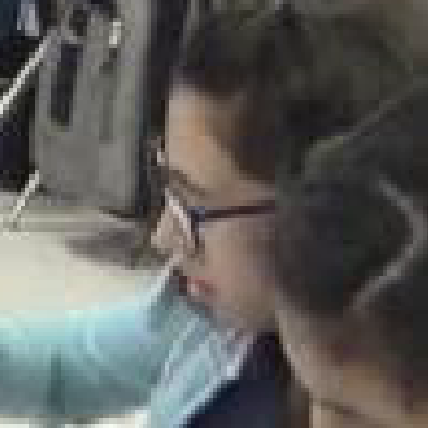}%
	}
	~
	\subfloat{\includegraphics[width=0.155\textwidth]{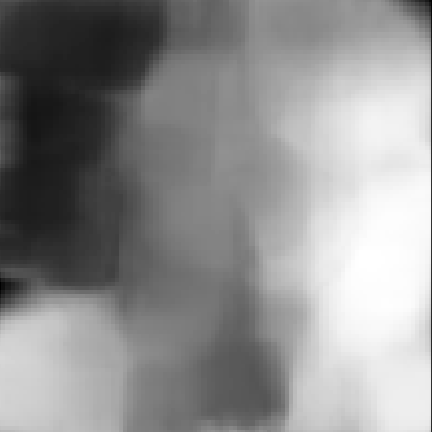}%
	}
	
	\caption{Examples of input video frames and the 3-second projection images.
		The top two rows show examples of talking video segments.
		The bottom two rows show examples of non-talking video segments.}
	\label{fig:sum_mag}
\end{figure}

For each head detection, we produce 3-second video clip proposals
   for detecting talking activities.
Over these regions, we compute dense optical flow estimates using
   Farneback's algorithm \cite{farneback2003two}.
At each pixel, we evaluate 
   $\log({\tt mag}(i,j)+0.01)$ where 
   ${\tt mag}(i,j)$ represents the magnitude
   of each motion vector.
Over each video segment, we then compute 
   the projection image as given by:   
   $$ P(i,j) \sum_{\text{all frames} f} \log ({\tt mag}_f (i,j)+0.01).
   $$   
We then train a variety of proposal
   region classifiers to differentiate between
   talking and non-talking activities.
   
We present example projection images in Fig \ref{fig:sum_mag}.
From the examples, compared to projections of talking activities,
   it is clear that projections of
   non-talking activities are characterized by dark regions
   around the mouth regions.   
   
For classifying the projected images, we consider simple classifiers.
We considered a modified LeNet5, XGBoost, AdaBoost, 
   decision tree, K-NN, quadratic discriminant analysis, and random forest classifier. 
Over the training set, we select the best three performing classifiers
   based on accuracy, AUC score, and F1 score, and then use
   a simple majority vote to combine them into a single system.

\begin{figure}[!b]
	\subfloat{\includegraphics[width=0.49\textwidth]{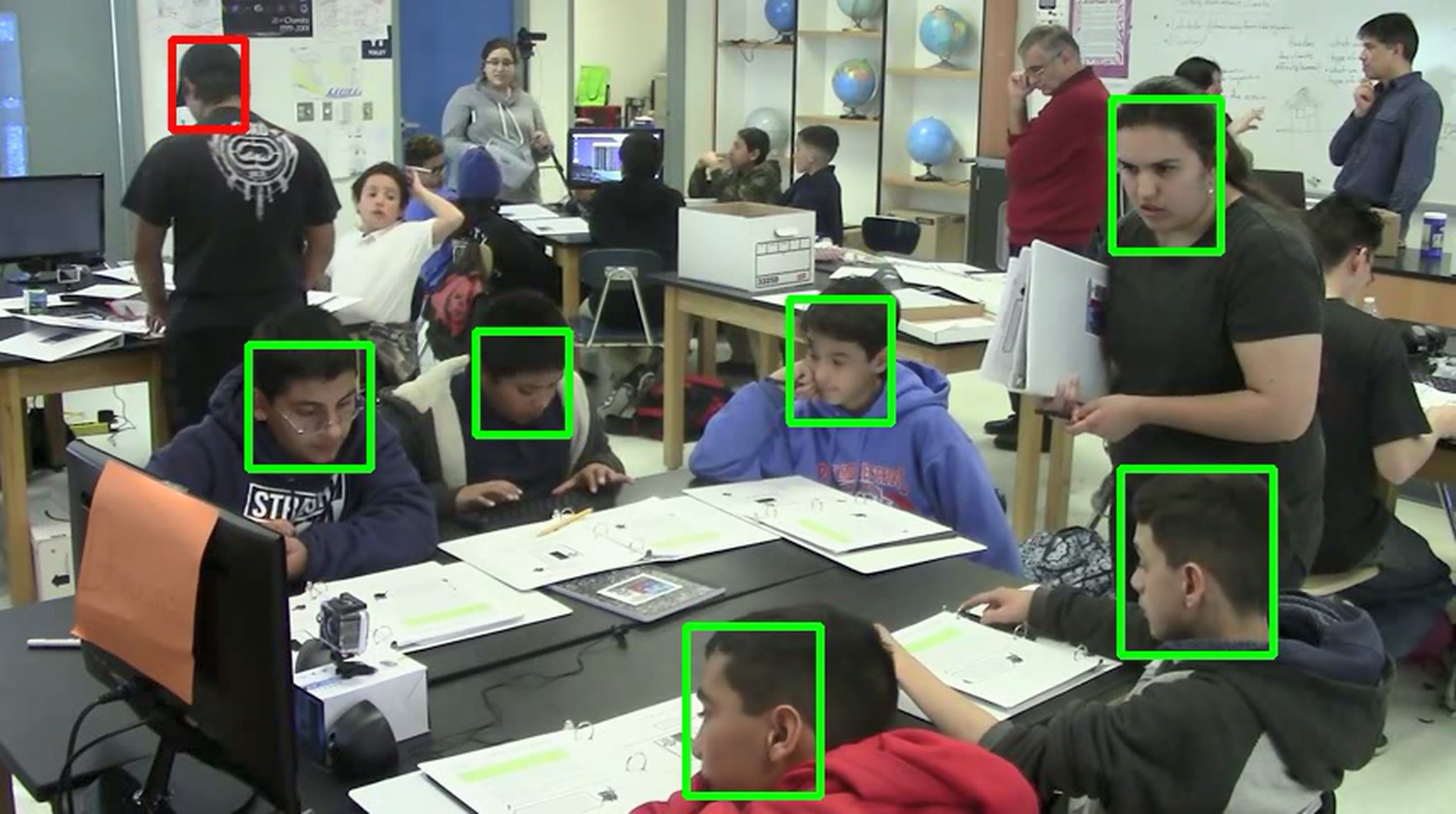}%
	}
	~
	\subfloat{\includegraphics[width=0.49\textwidth]{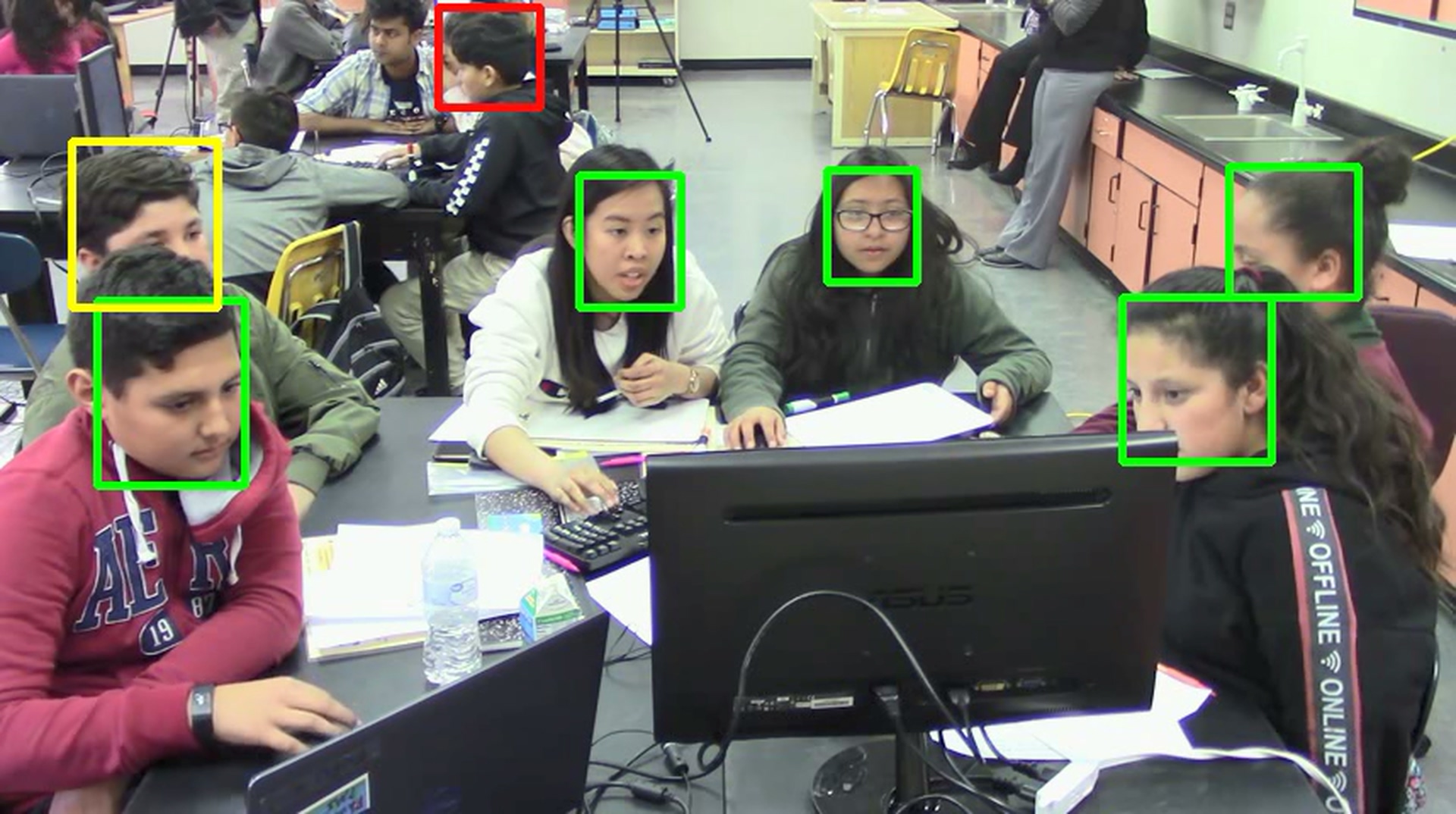}%
	}
	\caption{Results of head detection. 
		True positives are bounded by green boxes.
		False positives are bounded by red boxes.
		False negatives are bounded by yellow boxes.}
	\label{fig:HeadResults}
\end{figure}

\section{Results}\label{sec:results}
We summarize our results into three subsections.
First, we present results for our head detector
       in section \ref{sec:head}.
Second, we present results for
       our head region video detection results in
       section \ref{sec:roi}.
Third, we present final results for the 
       full system in section   
       \ref{sec:vid}.

\subsection{Head Detection System Results}\label{sec:head} 
We summarize head detection results 
    in Table  \ref{HeadResults}.
For training head detection, we selected 
   1,000 head examples
   and 1,200 non-face examples selected from 54 different video sessions.
We then tested our head detector on four 
   unseen videos as summarized
   in Table  \ref{HeadResults}.
We can see from the results that our proposed approach
   achieved F1 scores that range from 0.81 to 0.87  
   over 905,550 labeled students.

Example detections are shown in 
   Fig \ref{fig:HeadResults}.
We can see from the example that our 
   head classification
   system rejected all but one of the 
   distant detections.
Furthermore, we missed a single face due to occlusion.

\begin{table}[!t]
	\caption{\label{HeadResults}
		Results for student group detection over four videos.
		We present results over 905,550 labeled students.
		F1 scores are given for each video.   
		The videos represent different student groups.
		TP refers to true positives.
		FP refers to false positives.
		FN refers to false negatives.
	}
	\begin{center}
		\begin{tabular}{p{0.15\textwidth}p{0.15\textwidth}p{0.15\textwidth}p{0.12\textwidth}p{0.12\textwidth}p{0.12\textwidth}p{0.12\textwidth}}\toprule
			\textbf{Video}                 & \textbf{Labeled Students} & \textbf{Detected Students} & \textbf{TP} & \textbf{FP} & \textbf{FN} & \textbf{F1} \\ \toprule
			\textbf{V1} & 242,700                    & 180,640                     & 169,550      & 11,090       & 69,190       & 0.81        \\
			\textbf{V2} & 131,100                    & 122,230                     & 107,360      & 14,870       & 17,360       & 0.87        \\
			\textbf{V3} & 277,830                    & 229,810                     & 207,230      & 22,580       & 60,270       & 0.83        \\
			\textbf{V4} & 253,920                    & 230,600                     & 206,860      & 23,740       & 35,750       & 0.87        \\ \bottomrule
		\end{tabular}
	\end{center}
\end{table}

\begin{table}[!b]
\caption{\label{TrainingDataset}
                       Training dataset for talking detection. 
	Video names include the cohort number and the level number (e.g., C3L1).}
\begin{center}
\begin{tabular}{p{0.12\textwidth}p{0.15\textwidth}p{0.12\textwidth}p{0.15\textwidth}p{0.15\textwidth}p{0.15\textwidth}}
\toprule
\textbf{Group ID} & \textbf{Cohort} & \textbf{Group} & \textbf{Date} & \textbf{Urban/ Rural} & \textbf{Frame Rate (fps)} \\ \midrule
1        & C1L1   & D     & May-04 & Rural       & 60               \\
2        & C1L1   & D     & May-11 & Rural       & 60               \\
3        & C1L1   & C     & May-02 & Urban       & 60               \\
4        & C1L1   & C     & May-09 & Urban       & 60               \\
5        & C1L2   & A     & Jun-22 & Rural       & 60               \\
6        & C2L1   & A     & Mar-22 & Rural       & 30               \\
7        & C2L1   & A     & Apr-19 & Rural       & 30               \\
8        & C2L1   & A     & May-05 & Rural       & 30               \\
9        & C2L1   & A     & May-10 & Rural       & 30               \\
10       & C2L1   & B     & Mar-22 & Rural       & 30               \\
11       & C2L1   & D     & Feb-23 & Rural       & 30               \\
12       & C2L1   & D     & Mar-22 & Rural       & 30               \\
13       & C2L1   & A     & Feb-20 & Urban       & 30               \\ \bottomrule
\end{tabular}
\end{center}
\end{table}

\begin{table}[!t]
\caption{\label{ValDataset}
	Validation dataset for talking detection.
	Video names include the cohort number and the level number (e.g., C3L1).}
\begin{center}
\begin{tabular}{p{0.12\textwidth}p{0.15\textwidth}p{0.12\textwidth}p{0.15\textwidth}p{0.15\textwidth}p{0.15\textwidth}}
\toprule
\textbf{Group ID} & \textbf{Cohort} & \textbf{Group} & \textbf{Date} & \textbf{Urban/ Rural} & \textbf{Frame Rate (fps)} \\ \midrule
1        & C1L1   & B     & Mar-02 & Rural       & 30               \\
2        & C1L1   & C     & Mar-30 & Rural       & 60               \\
3        & C1L1   & C     & Apr-06 & Rural       & 60               \\
4        & C1L1   & C     & Apr-13 & Rural       & 60               \\
5        & C1L1   & E     & Mar-02 & Rural       & 60               \\
6        & C2L1   & B     & Feb-23 & Rural       & 30               \\
7        & C2L1   & C     & Apr-12 & Rural       & 30               \\
8        & C2L1   & D     & Mar-08 & Rural       & 30               \\
9        & C2L1   & E     & Apr-12 & Rural       & 30               \\
10       & C2L1   & B     & Feb-27 & Urban       & 30               \\
11       & C3L1   & C     & Apr-11 & Rural       & 30               \\
12       & C3L1   & D     & Feb-21 & Rural       & 30               \\
13       & C3L1   & D     & Mar-19 & Urban       & 30               \\ \bottomrule
\end{tabular}
\end{center}
\end{table}

\begin{table}[!b]
	\caption{\label{Results1}
		Head-based video region classification results.}
	\begin{center}
		\begin{tabular}{p{0.18\textwidth}p{0.15\textwidth}p{0.11\textwidth}p{0.15\textwidth}p{0.11\textwidth}p{0.11\textwidth}p{0.15\textwidth}}
			
			\toprule %
			\textbf{Methods}              & \textbf{Accuracy} & \textbf{\begin{tabular}[c]{@{}l@{}}AUC\\ Score\end{tabular}} & \textbf{Precision} & \textbf{Recall} & \textbf{F1} & \textbf{\begin{tabular}[c]{@{}l@{}}Confusion\\ Matrix\end{tabular}}                                                       \\ \midrule
			\textbf{LeNet5}               & 70\%              & 0.76               & 0.69               & 0.76            & 0.72        & $ \begin{bmatrix}1785 & 960\\    702 &  2177\end{bmatrix} $ \\~\\
			\textbf{XGBoost}              & 67\%              & 0.73               & 0.65               & 0.78            & 0.71        & $ \begin{bmatrix}1549 &1196\\    647  & 2232\end{bmatrix} $   \\~\\
			\textbf{AdaBoost}             & 70\%              & 0.70               & 0.64               & 0.72            & 0.67        & $ \begin{bmatrix}1557 &1188\\    810  & 2069\end{bmatrix} $   \\~\\
			\textbf{\begin{tabular}[c]{@{}l@{}}Decision \\Tree\end{tabular}}         & 59\%              & 0.59               & 0.60               & 0.60            & 0.60        & $ \begin{bmatrix}1598 &1147\\    1138& 1741\end{bmatrix} $    \\~\\
			\textbf{KNN}                  & 68\%              & 0.74               & 0.68               & 0.71            & 0.70        & $ \begin{bmatrix}1779 &966\\    831 &  2048\end{bmatrix} $    \\~\\
			\textbf{QDA}                  & 61\%              & 0.71               & 0.82               & 0.30            & 0.44        & $ \begin{bmatrix}2562 &183\\    2026 &853\end{bmatrix} $     \\~\\
			\textbf{\begin{tabular}[c]{@{}l@{}}Random \\Forest\end{tabular}}         & 62\%              & 0.65               & 0.61               & 0.75            & 0.67        & $ \begin{bmatrix}1354 &1391\\    728  & 2151\end{bmatrix} $   \\~\\
			\textbf{\begin{tabular}[c]{@{}l@{}}XGBoost+\\AdaBoost+\\ KNN\end{tabular}}& 79\%              & 0.77               & 0.69               & 0.72            & 0.70        & $ \begin{bmatrix}1810& 935\\    804 &  2075\end{bmatrix} $    \\
			
			\bottomrule
		\end{tabular}   
	\end{center}
\end{table}

\subsection{Head Video Region Classification Results}\label{sec:roi} 
In this section,
   we provide comparisons
   against single activity classifiers.
For this purpose, we crop
   head regions and resize them
   to $100 \times 100$ pixels.
For our comparisons, each video
   segment is clipped at 3 seconds.
We report results on two
   datasets.
The first dataset is used for
   selecting the classifiers
   that are used in our
   majority classification
   system.
We use a second dataset
   to assess the performance
   of the majority classifier
   on four videos that
   range from 11 to 24 minutes.    

\begin{table}[!t]
	\caption{\label{Results2}
		Talking detection for long videos.}
	\begin{center}
		\begin{tabular}{p{0.15\textwidth}p{0.15\textwidth}p{0.25\textwidth}p{0.18\textwidth}p{0.15\textwidth}p{0.15\textwidth}}
			\toprule
			\textbf{Video}  & \textbf{Duration}                  & \textbf{Person Label}           & \textbf{Ours} & \textbf{TSN} & \textbf{C3D} \\ \toprule
			\multirow{5}{*}{\textbf{V1}} & \multirow{5}{*}{23 min 45 s}&\textit{\textbf{Issac}}         & \textbf{66\%}                          & 28\%         & 28\%         \\
			& & \textit{\textbf{Julia7P}}       &   \textbf{48\%}                          & 38\%         & 33\%         \\
			& & \textit{\textbf{Martina64P}}  &    \textbf{58\%}                          &11\%         & 31\%         \\
			& & \textit{\textbf{Suzie66P}}   &    \textbf{44\%}                          & 11\%         & 7\%         \\
			& & \textit{\textbf{Bernard129P}} &   \textbf{51\%}                          & 18\%         & 19\%         \\ \cmidrule{3-6}
			& & \textbf{Average}   & \textbf{53\%}                          & 21\%         & 24\%         \\ \midrule
			\multirow{6}{*}{\textbf{V2}} & \multirow{6}{*}{11 min 20 s}& \textit{\textbf{Irma}}          & 53\%                       &\textbf{67\%}              &64\%        \\
			&& \textit{\textbf{Emilio25P}}     & \textbf{68\%}                          &21\%              &14\%              \\
			&& \textit{\textbf{Herminio10P}}   & 56\%                          &72\%              &\textbf{79\%}              \\
			&& \textit{\textbf{Jacinto51P}}    & \textbf{66\%}                          &21\%              &41\%              \\
			&& \textit{\textbf{Jorge17P}}      & \textbf{60\%}                          &53\%              &43\%              \\
			&& \textit{\textbf{Juan16P}}       & \textbf{62\%}                          &39\%              &35\%              \\ \cmidrule{3-6}
			&& \textbf{Average}   & \textbf{61\%}                          & 46\%         & 46\%         \\ \midrule
			
			\multirow{6}{*}{\textbf{V3}} & \multirow{6}{*}{16 min 6 s} & \textit{\textbf{Kelly}}         & 70\%               &67\%              &\textbf{71\%}              \\
			&& \textit{\textbf{Marta12P}}      & \textbf{68\%}                          & 19\%             &34\%             \\
			&& \textit{\textbf{Cindy14P}}      & \textbf{74\%}                          & 23\%             &\textbf{74\%}              \\
			&& \textit{\textbf{Carmen13P}}     & \textbf{51\%}                          &31\%              &50\%              \\
			&& \textit{\textbf{Marina15P}}     & \textbf{64\%}                          &22\%              &26\%              \\
			&& \textit{\textbf{Scott}}         & 87\%                          &\textbf{95\%}              &92\%              \\ \cmidrule{3-6}
			&& \textbf{Average}   & \textbf{69\%}                          & 43\%         & 58\%         \\ \midrule
			
			\multirow{6}{*}{\textbf{V4}} & \multirow{6}{*}{23 min 45 s}& \textit{\textbf{Phuong}}        & 58\%              &\textbf{71\%}              &58\%              \\
			&& \textit{\textbf{Jacob103P}}     & \textbf{53\%}                          &51\%              &46\%              \\
			&& \textit{\textbf{Josephina104P}} & 42\%                          &\textbf{63\%}              &47\%              \\
			&& \textit{\textbf{Juanita107P}}   & 55\%                          &\textbf{64\%}              &60\%              \\
			&& \textit{\textbf{Tina105P}}      & \textbf{55\%}                          &44\%              &47\%             \\
			&& \textit{\textbf{Vincent106P}}   & \textbf{45\%}                          &43\%              &40\%              \\\cmidrule{3-6}
			&& \textbf{Average}   & 51\%                          & \textbf{56\%}         & 50\%         \\ \midrule
			&&\textbf{Overall Average}                             & \textbf{59\%}         & 42\% &45\%        \\
			
			\bottomrule
		\end{tabular}
	\end{center}
\end{table}

For training the proposed classification
    method and all other methods, 
    we use 11,315 video clips
   extracted from 13 different video sessions,
   with a total of 27 students (see Table \ref{TrainingDataset}).
For the validation set,
    we use 5,624 video clips
    extracted from 13  video sessions,
    with a total of 37 students (see Table \ref{ValDataset}).
Table \ref{Results1} summarizes
    the results from using different classifiers.
We chose XGBoost, AdaBoost, and KNN for the 
    voting system.
Over our validation set, this combination
    gave the highest accuracy at 79\%.    
For comparing our system against alternative 
    approaches, we use four different videos
    as summarized in Table \ref{Results2}.
From the results, our system gave an average
    accuracy of 59\% compared to
    42\% for TSN and 45\% for C3D.

\begin{figure}[!t]
\subfloat{\includegraphics[width=0.49\textwidth]{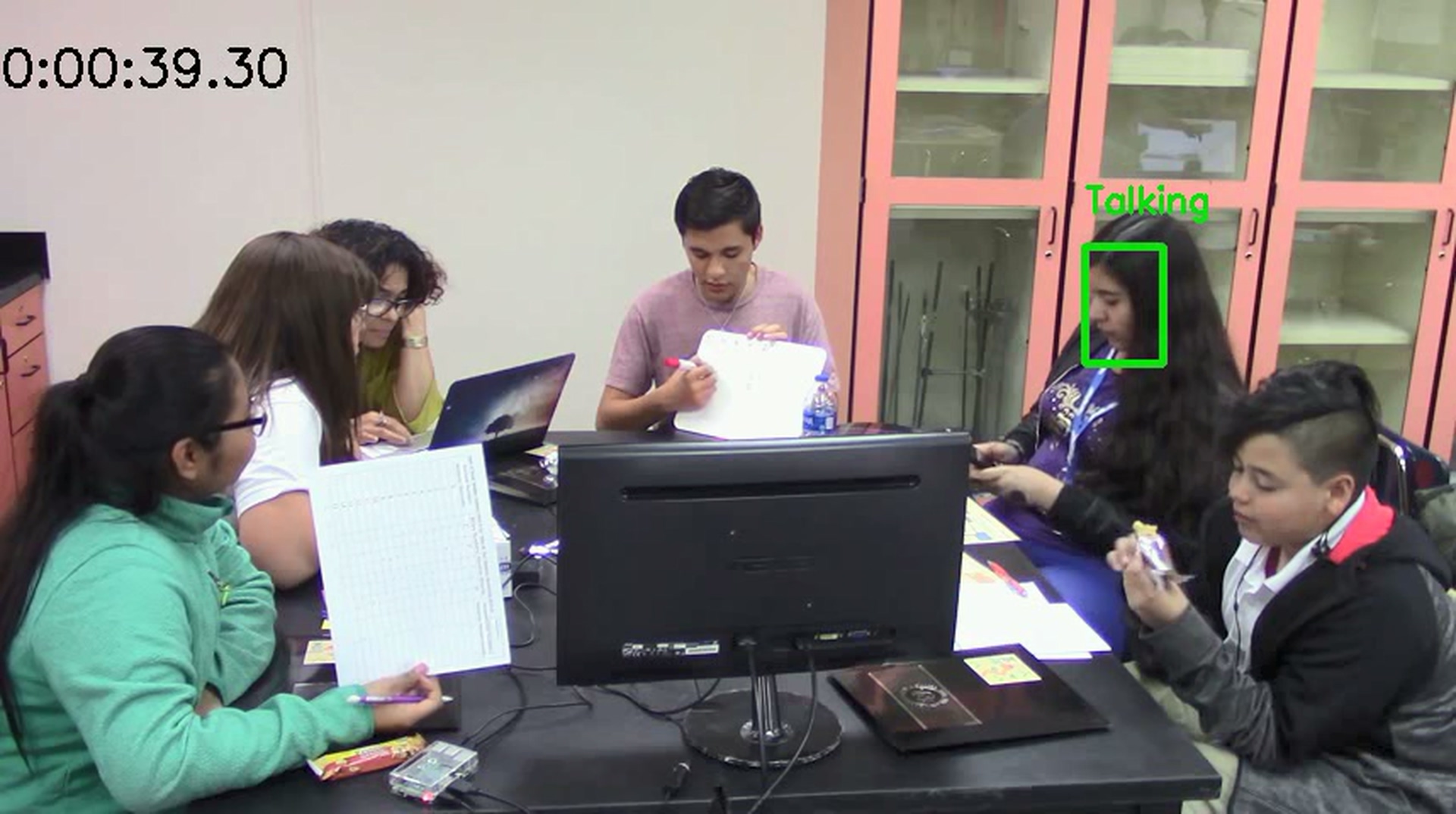}%
}
~
\subfloat{\includegraphics[width=0.49\textwidth]{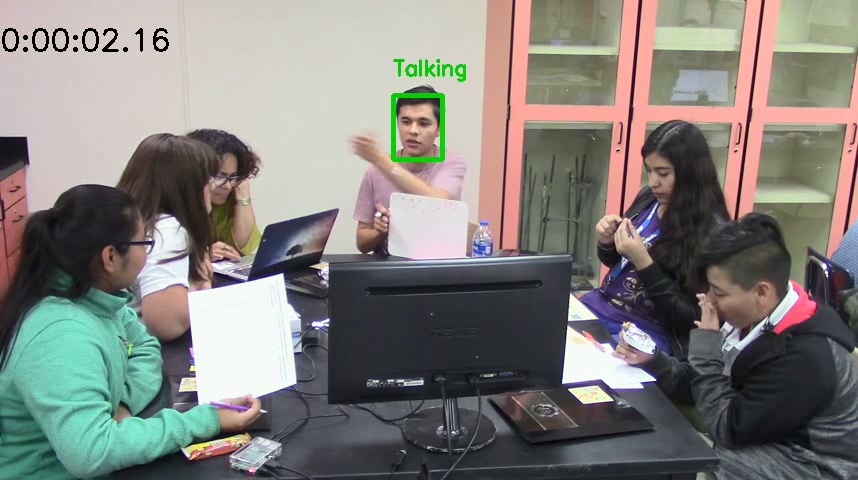}%
}
\\
\subfloat{\includegraphics[width=0.49\textwidth]{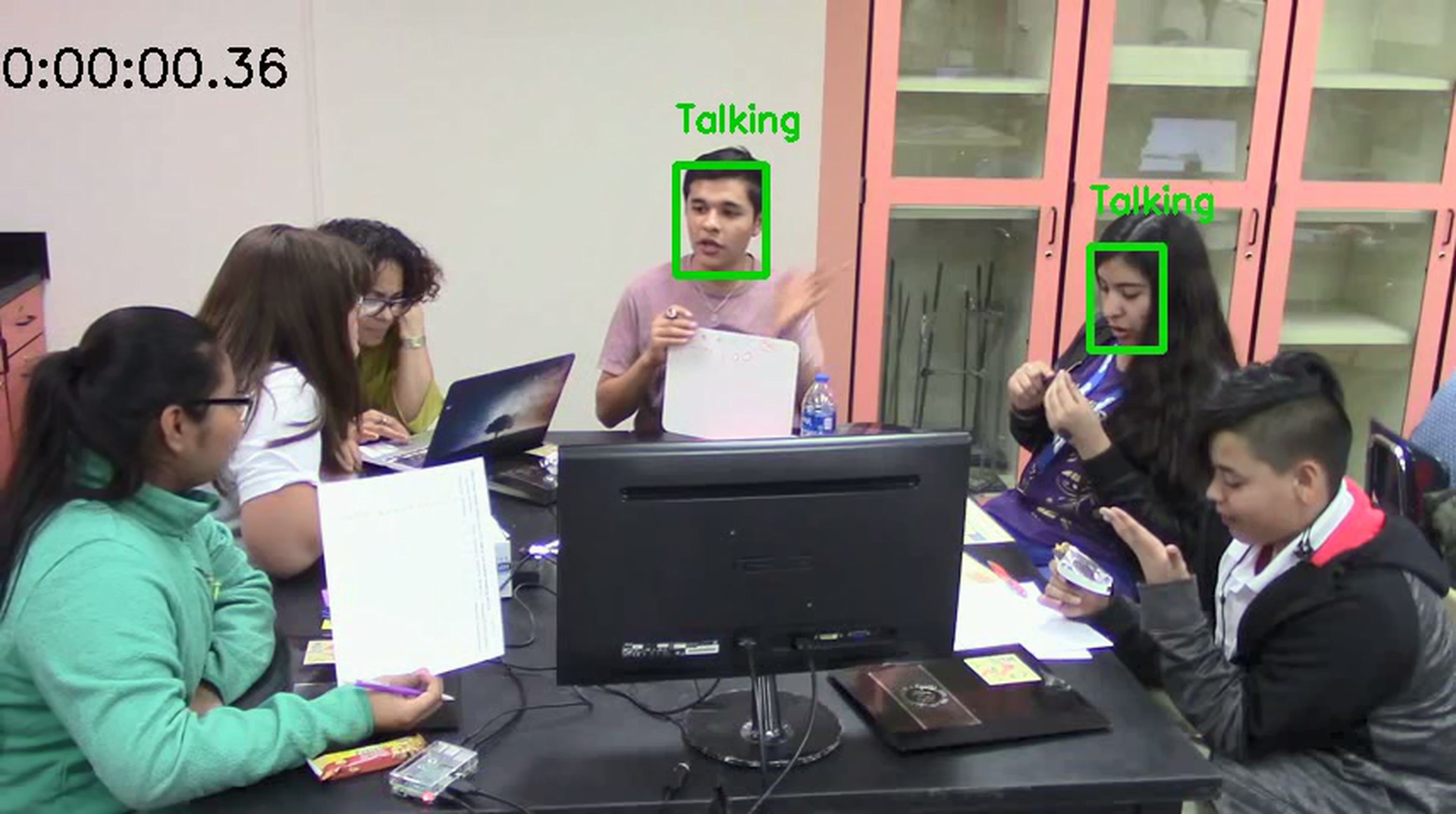}%
}
~
\subfloat{\includegraphics[width=0.49\textwidth]{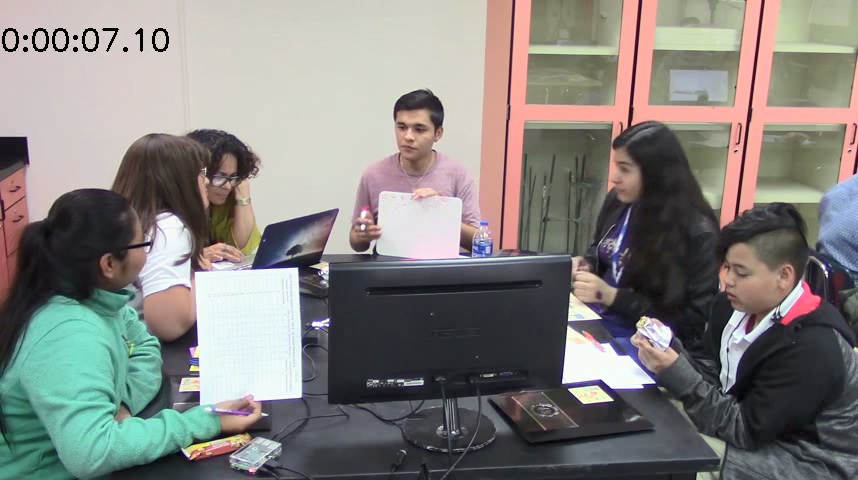}%
}
\caption{Example of talking detection on the original video.}
\label{fig:TalkingVideoResults}
\end{figure}

\subsection{Talking Activity Detection System}\label{sec:vid}
We present an example of the final
   system in Fig. \ref{fig:TalkingVideoResults}.
As shown in Fig. \ref{fig:TalkingVideoResults},
   our system detects who is talking and
   places a bounding box identifying the person
   talking.
Furthermore, unlike single activity
   systems like TSN and C3D, we can
   detect multiple people talking at the same time.

\section{Conclusion}\label{sec:conclusion}
We presented a new method for detecting
   students talking in collaborative
   learning environment videos.
Our approach combines head detection
   with activity detection using
   a projection of motion vectors and
   a majority voting classification system.
Our approach significantly outperformed
   single activity classification systems.
Yet, our average accuracy at 59\% suggests
   that there is still room for significant
   improvement.
Our approach will also need to be
   further validated before
   it can be effectively used by educational
   researchers.


%
%
%
\bibliographystyle{splncs04}
\bibliography{bare_conf}
\end{document}